\documentclass[11pt,a4paper]{article}
\usepackage[hyperref]{acl2019}
\usepackage{times}
\usepackage{latexsym}

\usepackage{url}

\usepackage{xspace}
\usepackage{amsmath}
\usepackage{amsfonts}
\usepackage{array,multirow}
\usepackage{pgfplots}
\usepackage{tikz}
\usepackage{subfig}
\usepackage{array} % set tabular width
\usepackage{verbatim}

\usepackage{graphicx} %插入图片的宏包

\usetikzlibrary{backgrounds,fit}
\usetikzlibrary{shapes,arrows,shadows}
\usetikzlibrary{patterns}
\usetikzlibrary{shapes.geometric}
\usetikzlibrary{decorations.pathreplacing}
\usetikzlibrary{calc}
\usepackage{arydshln}
\usepackage{color}
\usepackage{caption}
\usepackage{booktabs}
\usepackage{cleveref}
\crefname{section}{§}{§§}
\Crefname{section}{§}{§§}
\usepackage{enumitem}

\newcommand{\PreserveBackslash}[1]{\let\temp=\\#1\let\\=\temp}
\newcolumntype{C}[1]{>{\PreserveBackslash\centering}p{#1}}
\newcolumntype{R}[1]{>{\PreserveBackslash\raggedleft}p{#1}}
\newcolumntype{L}[1]{>{\PreserveBackslash\raggedright}p{#1}}
\newcolumntype{M}[1]{ >{\centering\arraybackslash}m{#1}}

\newlength{\vseg}
\newlength{\hseg}
\newlength{\wnode}
\newlength{\hnode}
\newcommand{\name}{{\normalsize{D}}{\small{LCL}}\xspace}
\newcommand{\plainname}{DLCL\xspace}

\definecolor{ugreen}{rgb}{0,0.5,0}
\definecolor{lgreen}{rgb}{0.9,1,0.8}
\definecolor{lightgray}{gray}{0.85}
\usepackage{xcolor}
\definecolor{myblack}{rgb}{0.15,0.15,0.15}
\definecolor{lyellow}{rgb}{0.54, 0.25, 0.27}

\aclfinalcopy % Uncomment this line for the final submission
\def\aclpaperid{362} %  Enter the acl Paper ID here

\title{Learning Deep Transformer Models for Machine Translation}

\author{
	Qiang Wang$^1$,
	Bei Li$^1$,
	Tong Xiao$^{1,2}$\thanks{\xspace\xspace Corresponding author.},
	Jingbo Zhu$^{1,2}$,
	\textbf{Changliang Li$^{3}$,} \\
	\textbf{Derek F. Wong$^{4}$,}
	\textbf{Lidia S. Chao$^{4}$} \\
	$^{1}$NLP Lab, Northeastern University, Shenyang, China\\
	$^{2}$NiuTrans Co., Ltd., Shenyang, China \\
	$^{3}$Kingsoft AI Lab, Beijing, China \\
	$^{4}$NLP$^2$CT Lab, University of Macau, Macau, China \\
	{\tt
		wangqiangneu@gmail.com, libei\_neu@outlook.com,
		}\\
	{\tt
		\{xiaotong,zhujingbo\}@mail.neu.edu.com,
		} \\
	{\tt
		lichangliang@kingsoft.com,
		\{derekfw,lidiasc\}@um.edu.mo
	} \\
}

\date{}

\begin{document}
\maketitle
\begin{abstract}
    Transformer is the state-of-the-art model in recent machine translation evaluations. Two strands of research are promising to improve models of this kind: the first uses wide networks (a.k.a. Transformer-Big) and has been the de facto standard for the development of the Transformer system, and the other uses deeper language representation but faces the difficulty arising from learning deep networks. Here, we continue the line of research on the latter. We claim that a truly deep Transformer model can surpass the Transformer-Big counterpart by 1) proper use of layer normalization and 2) a novel way of passing the combination of previous layers to the next. On WMT'16 English-German, NIST OpenMT'12 Chinese-English and larger WMT'18 Chinese-English tasks, our deep system (30/25-layer encoder)  outperforms the shallow Transformer-Big/Base baseline (6-layer encoder) by 0.4$\sim$2.4 BLEU points. As another bonus, the deep model is 1.6X smaller in size and 3X faster in training than Transformer-Big\footnote{The source code is available at \url{https://github.com/wangqiangneu/dlcl}}.
\end{abstract}

\section{Introduction}

\noindent Neural machine translation (NMT) models have advanced the previous state-of-the-art by learning mappings between sequences via neural networks and attention mechanisms \cite{sutskever2014sequence,bahdanau2014neural}. The earliest of these read and generate word sequences using a series of recurrent neural network (RNN) units, and the improvement continues when 4-8 layers are stacked for a deeper model \cite{luong2015effective,wu2016google}. More recently, the system based on multi-layer self-attention (call it Transformer) has shown strong results on several large-scale tasks \cite{vaswani2017attention}. In particular, approaches of this kind benefit greatly from a wide network with more hidden states (a.k.a. Transformer-Big), whereas simply deepening the network has not been found to outperform the ``shallow'' counterpart \cite{bapna2018training}. Do deep models help Transformer? It is still an open question for the discipline.

For vanilla Transformer, learning deeper networks is not easy because there is already a relatively deep model in use\footnote{For example, a standard Transformer encoder has 6 layers. Each of them consists of two sub-layers. More sub-layers are involved on the decoder side.}. It is well known that such deep networks are difficult to optimize due to the gradient vanishing/exploding problem \cite{pascanu2013difficulty,bapna2018training}. We note that, despite the significant development effort, simply stacking more layers cannot benefit the system and leads to a disaster of training in some of our experiments.

A promising attempt to address this issue is \citet{bapna2018training}'s work. They trained a 16-layer Transformer encoder by using an enhanced attention model. In this work, we continue the line of research and go towards a much deeper encoder for Transformer. We choose encoders to study because they have a greater impact on performance than decoders and require less computational cost \cite{domhan2018much}. Our contributions are threefold:

\begin{itemize}
	\item We show that the proper use of layer normalization is the key to learning deep encoders. The deep network of the encoder can be optimized smoothly by relocating the layer normalization unit. While the location of layer normalization has been discussed in recent systems \cite{vaswani2018tensor2tensor,domhan2018much,klein2017opennmt}, as far as we know, its impact has not been studied in deep Transformer.
	\item Inspired by the linear multi-step method in numerical analysis \cite{ascher1998computer}, we propose an approach based on dynamic linear combination of layers (\name) to memorizing the features extracted from all preceding layers. This overcomes the problem with the standard residual network where a residual connection just relies on the output of one-layer ahead and may forget the earlier layers.
	\item We successfully train a 30-layer encoder, far surpassing the deepest encoder reported so far \cite{bapna2018training}. To our best knowledge, this is the deepest encoder used in NMT.
\end{itemize}

On WMT'16 English-German, NIST OpenMT'12 Chinese-English, and larger WMT'18 Chinese-English translation tasks, we show that our deep system (30/25-layer encoder) yields a BLEU improvement of 1.3$\sim$2.4 points over the base model (Transformer-Base with 6 layers). It even outperforms Transformer-Big by 0.4$\sim$0.6 BLEU points, but requires 1.6X fewer model parameters and 3X less training time. More interestingly, our deep model is 10\% faster than Transformer-Big in inference speed.

\section{Post-Norm and Pre-Norm Transformer}

\begin{figure}[t]
	\begin{center}
		\renewcommand{\arraystretch}{0}
		\begin{tabular}{C{.49\textwidth}}
			\subfloat[post-norm residual unit]
			{
				\begin{tikzpicture}{baseline}
				\begin{scope}
				
				\setlength{\vseg}{1.5em}
				\setlength{\hseg}{2.5em}
				\setlength{\wnode}{3em}
				\setlength{\hnode}{2em}
				
				\tikzstyle{layernode} = [draw, thin, rounded corners=1pt, inner sep=3pt, fill=yellow!20,  minimum width=0.7\wnode, minimum height=0.3\hnode]
				
				% input
				\node [] (input) at (0,0) {$x_l$};
				
				% module
				\node[layernode, anchor=west, fill=red!15] (module) at ([xshift=\vseg]input.east) {$\mathcal{F}$};
				
				% residual
				\node[ inner sep=0pt, anchor=west, minimum width=0.15\wnode, minimum height=0.1\hnode] (add1) at ([xshift=\vseg]module.east) {$\bigoplus$};
				
				% module
				\node[layernode, anchor=west,  fill=red!15] (ln) at ([xshift=\vseg]add1.east) {LN};
				
				% output
				\node [] (output) at ([xshift=2\vseg]ln.east) {$x_{l+1}$};
				
				\draw[->, thick] (input.east) to node [auto] {} (module.west);
				\draw[->, thick] (module.east) to node [auto] {} (add1.west);
				\draw[->, thick] (add1.east) to node [auto] {$y_l$} (ln.west);
				\draw[->, thick] (ln.east) to node [auto] {} (output.west);
				
				\draw [->, thick ,rounded corners] (input.north) -- ([yshift=0.9em]input.north)  -- ([yshift=1em]add1.north) -- ([yshift=1pt]add1.north);
				
				\end{scope}
				\label{fig:post}
				\end{tikzpicture}
			}
			\\
			\subfloat[pre-norm residual unit]
			{
				\begin{tikzpicture}{baseline}
				\begin{scope}
				
				\setlength{\vseg}{1.5em}
				\setlength{\hseg}{2.5em}
				\setlength{\wnode}{3em}
				\setlength{\hnode}{2em}
				
				\tikzstyle{layernode} = [draw, thin, rounded corners=1pt, inner sep=3pt, fill=yellow!20,  minimum width=0.7\wnode, minimum height=0.3\hnode]
				
				% input
				\node [] (input) at (0,0) {$x_l$};
				
				% ln
				\node[layernode, anchor=west, fill=red!15] (ln) at ([xshift=\vseg]input.east) {LN};
				
				% module
				\node[layernode, anchor=west,  fill=red!15] (module) at ([xshift=\vseg]ln.east) {$\mathcal{F}$};
				
				% residual
				\node[ inner sep=0pt, anchor=west, minimum width=0.15\wnode, minimum height=0.1\hnode] (add1) at ([xshift=\vseg]module.east) {$\bigoplus$};
				
				% output
				\node [] (output) at ([xshift=2\vseg]add1.east) {$x_{l+1}$};
				
				\draw[->, thick] (input.east) to node [auto] {} (ln.west);
				\draw[->, thick] (ln.east) to node [auto] {} (module.west);
				\draw[->, thick] (module.east) to node [auto] {$y_l$} (add1.west);
				\draw[->, thick] (add1.east) to node [auto] {} (output.west);
				
				\draw [->, thick ,rounded corners] (input.north) -- ([yshift=0.9em]input.north)  -- ([yshift=1em]add1.north) -- ([yshift=1pt]add1.north);
				
				\end{scope}
				\label{fig:prev}
				\end{tikzpicture}
			}	\\
		\end{tabular}
		
	\end{center}
	
	\begin{center}
		\vspace{-0.5em}
		\caption{Examples of pre-norm residual unit and post-norm residual unit. $\mathcal{F}$ = sub-layer, and LN = layer normalization.}
		\label{fig:post-prev}
		\vspace{-1.5em}
	\end{center}
\end{figure}

\noindent The Transformer system and its variants follow the standard encoder-decoder paradigm.
On the encoder side, there are a number of identical stacked layers. Each of them is composed of a self-attention sub-layer and a feed-forward sub-layer. The attention model used in Transformer is multi-head attention, and its output is fed into a fully connected feed-forward network.
Likewise, the decoder has another stack of identical layers. It has an encoder-decoder attention sub-layer in addition to the two sub-layers used in each encoder layer. In general, because the encoder and the decoder share a similar architecture, we can use the same method to improve them. In the section, we discuss a more general case, not limited to the encoder or the decoder.

\subsection{Model Layout}
\noindent  For Transformer, it is not easy to train stacked layers on neither the encoder-side nor the decoder-side. Stacking all these sub-layers prevents the efficient information flow through the network, and probably leads to the failure of training. Residual connections and layer normalization are adopted for a solution. Let $\mathcal{F}$ be a sub-layer in encoder or decoder, and $\theta_l$ be the parameters of the sub-layer. A residual unit is defined to be \cite{he2016identity}:
\begin{eqnarray}
\label{eq:x} x_{l+1} &=& f(y_l) \\
\label{eq:y} y_l &=& x_l + \mathcal{F}(x_l; \theta_l)
\end{eqnarray}

\noindent where $x_l$ and $x_{l+1}$ are the input and output of the $l$-th sub-layer, and $y_l$ is the intermediate output followed by the post-processing function $f(\cdot)$. In this way, $x_l$ is explicitly exposed to $y_l$ (see Eq. (\ref{eq:y})).

Moreover, layer normalization is adopted to reduce the variance of sub-layer output because hidden state dynamics occasionally causes a much longer training time for convergence. There are two ways to incorporate layer normalization into the residual network.

\begin{itemize}
	\item \textbf{Post-Norm}. In early versions of Transformer \cite{vaswani2017attention}, layer normalization is placed after the element-wise residual addition (see Figure~\ref{fig:post-prev}(a)), like this:
	
	\begin{equation}
	\label{eq:post}
	x_{l+1} = \textrm{LN}(x_l + \mathcal{F}(x_l; \theta_l))
	\end{equation}
	
	where $\textrm{LN}(\cdot)$ is the layer normalization function, whose parameter is dropped for simplicity. It can be seen as a post-processing step of the output (i.e., $f(x)=\textrm{LN}(x)$).

	\item \textbf{Pre-Norm}. In recent implementations \cite{klein2017opennmt,vaswani2018tensor2tensor,domhan2018much}, layer normalization is applied to the input of every sub-layer (see Figure~\ref{fig:post-prev}(b)):
	
	\begin{equation}
	\label{eq:prev}
	x_{l+1} = x_l + \mathcal{F}(\textrm{LN}(x_l); \theta_l)
	\end{equation}
	
	Eq. (\ref{eq:prev}) regards layer normalization as a part of the sub-layer, and does nothing for post-processing of the residual connection (i.e., $f(x)=x$).\footnote{We need to add an additional function of layer normalization to the top layer to prevent the excessively increased value caused by the sum of unnormalized output.}
\end{itemize}

Both of these methods are good choices for implementation of Transformer. In our experiments, they show comparable performance in BLEU for a system based on a 6-layer encoder (Section \cref{sec:en2de_result}).

\subsection{On the Importance of Pre-Norm for Deep Residual Network}
\noindent
The situation is quite different when we switch to deeper models. More specifically, we find that pre-norm is more efficient for training than post-norm if the model goes deeper. This can be explained by seeing back-propagation which is the core process to obtain gradients for parameter update. Here we take a stack of $L$ sub-layers as an example. Let $\mathcal{E}$ be the loss used to measure how many errors occur in system prediction, and $x_L$ be the output of the topmost sub-layer. For post-norm Transformer, given a sub-layer $l$, the differential of $\mathcal{E}$ with respect to $x_l$ can be computed by the chain rule, and we have
\begin{eqnarray}
\frac{\partial{\mathcal{E}}}{\partial{x_{l}}} & = & \frac{\partial{\mathcal{E}}}{\partial{x_L}} \times \prod_{k=l}^{L-1} \frac{\partial \textrm{LN}(y_k)}{\partial y_k} \times \nonumber \\
\label{eq:post-bp} & & \prod_{k=l}^{L-1} \Big(1 + \frac{\partial{\mathcal{F}(x_k; \theta_k)}}{\partial{x_k}}\Big)
\end{eqnarray}

\noindent where $\prod_{k=l}^{L-1} \frac{\partial \textrm{LN}(y_k)}{\partial y_k}$ means the backward pass of the layer normalization, and $\prod_{k=l}^{L-1} (1 + \frac{\partial{\mathcal{F}(x_k; \theta_k)}}{\partial{x_k}})$ means the backward pass of the sub-layer with the residual connection. Likewise, we have the gradient for pre-norm
\footnote{For a detailed derivation, we refer the reader to Appendix A.}:
%\footnote{A detailed derivation can be found in Appendix A.}:
\begin{equation}
\label{eq:prev-bp}
\begin{split}
\frac{\partial{\mathcal{E}}}{\partial{x_{l}}} &= \frac{\partial{\mathcal{E}}}{\partial{x_L}} \times \Big(1 + \sum_{k=l}^{L-1} \frac{\partial{\mathcal{F}(\textrm{LN}(x_k); \theta_k)}}{\partial{x_l}}\Big)
\end{split}
\end{equation}

Obviously, Eq. (\ref{eq:prev-bp}) establishes a direct way to pass error gradient $\frac{\partial{\mathcal{E}}}{\partial{x_L}}$ from top to bottom. Its merit lies in that the number of product items on the right side does not depend on the depth of the stack.

\noindent In contrast, Eq. (\ref{eq:post-bp}) is inefficient for passing gradients back because the residual connection is not a bypass of the layer normalization unit (see Figure \ref{fig:post-prev}(a)). Instead, gradients have to be passed through $\textrm{LN}(\cdot)$ of each sub-layer. It in turn introduces term $\prod_{k=l}^{L-1} \frac{\partial \textrm{LN}(y_k)}{\partial y_k}$ into the right hand side of Eq. (\ref{eq:post-bp}), and poses a higher risk of gradient vanishing or exploring if $L$ goes larger. This was confirmed by our experiments in which we successfully trained a pre-norm Transformer system with a 20-layer encoder on the WMT English-German task, whereas the post-norm Transformer system failed to train for a deeper encoder (Section \cref{sec:en2de_result}).

\section{Dynamic Linear Combination of Layers}

%%%%%%%%%%%%%%%%
%% FIGURE STARTS
%% FIGURE: weights for different methods
\begin{figure*}[t]
	\begin{center}
		\setlength{\tabcolsep}{0pt}

		\begin{tabular}{C{.24\textwidth}C{.24\textwidth}C{.24\textwidth}C{.24\textwidth}}

			\subfloat [%\small{Residual Connection}
			] {
				\begin{tikzpicture}
				\begin{scope}	
				
				\setlength{\hnode}{1.2em}
				\tikzstyle{elementnode} = [rectangle,anchor=center]
				\tikzstyle{ynode} = [font=\small]
				\tikzstyle{xnode} = [left,font=\small,anchor=east]
				
				% weight
				\foreach \i / \j / \c in
				{0/3/1,
					0/2/0,   1/2/1,
					0/1/0,  1/1/0, 2/1/1,
					0/0/0,   1/0/0,  2/0/0, 3/0/1}
				\node[elementnode,minimum size=1*\hnode,inner sep=0.0pt, draw] (a\i\j) at (\hnode*\i, \hnode*\j) {\scriptsize{\c}};
				
				% x
				\node[xnode] (x1) at (a03.west) {$x_1$};
				\node[xnode] (x2) at (a02.west) {$x_2$};
				\node[xnode] (x3) at (a01.west) {$x_3$};
				\node[xnode] (x4) at (a00.west) {$x_4$};
				
				% y
				\node[ynode,anchor=south] (y0) at (a03.north) {$y_0$};
				\node[ynode] (y1) at ([xshift=0.45*\hnode, anchor=west]y0.east) {$y_1$};
				\node[ynode] (y2) at ([xshift=0.45*\hnode]y1.east) {$y_2$};
				\node[ynode] (y3) at ([xshift=0.45*\hnode]y2.east) {$y_3$};
				
				\end{scope}
				
				\end{tikzpicture}
			}
			&
			\subfloat [%\small{Dense Residual Connection}
			] {
				
				\begin{tikzpicture}
				\begin{scope}	
				
				\setlength{\hnode}{1.2em}
				\tikzstyle{elementnode} = [rectangle,anchor=center]
				\tikzstyle{ynode} = [font=\small]
				\tikzstyle{xnode} = [left,font=\small,anchor=east]
				
				% weight
				\foreach \i / \j / \c in
				{0/3/1,
					0/2/1,   1/2/1,
					0/1/1,  1/1/1, 2/1/1,
					0/0/1,   1/0/1,  2/0/1, 3/0/1}
				\node[elementnode,minimum size=1*\hnode,inner sep=0.0pt, draw] (a\i\j) at (\hnode*\i, \hnode*\j) {\scriptsize{\c}};
				
				% x
				\node[xnode] (x1) at (a03.west) {$x_1$};
				\node[xnode] (x2) at (a02.west) {$x_2$};
				\node[xnode] (x3) at (a01.west) {$x_3$};
				\node[xnode] (x4) at (a00.west) {$x_4$};
				
				% y
				\node[ynode,anchor=south] (y0) at (a03.north) {$y_0$};
				\node[ynode] (y1) at ([xshift=0.45*\hnode, anchor=west]y0.east) {$y_1$};
				\node[ynode] (y2) at ([xshift=0.45*\hnode]y1.east) {$y_2$};
				\node[ynode] (y3) at ([xshift=0.45*\hnode]y2.east) {$y_3$};
				
				\end{scope}
				\vspace{-0.5em}
				\end{tikzpicture}
			}
			&
			\subfloat [%\small{\citet{wang2018multi,bapna2018training}}
			] {
				
				\begin{tikzpicture}
				\begin{scope}	
				
				\setlength{\hnode}{1.2em}
				\tikzstyle{elementnode} = [rectangle,anchor=center]
				\tikzstyle{ynode} = [font=\small]
				\tikzstyle{xnode} = [left,font=\small,anchor=east]
				
				% weight
				\foreach \i / \j / \c in
				{0/3/1,
					0/2/0,   1/2/1,
					0/1/0,  1/1/0, 2/1/1,
					0/0/.1,   1/0/.3,  2/0/.2, 3/0/.4}
				\node[elementnode,minimum size=1*\hnode,inner sep=0.1pt, draw] (a\i\j) at (\hnode*\i, \hnode*\j) {\scriptsize{\c}};
				
				% x
				\node[xnode] (x1) at (a03.west) {$x_1$};
				\node[xnode] (x2) at (a02.west) {$x_2$};
				\node[xnode] (x3) at (a01.west) {$x_3$};
				\node[xnode] (x4) at (a00.west) {$x_4$};
				
				% y
				\node[ynode,anchor=south] (y0) at (a03.north) {$y_0$};
				\node[ynode] (y1) at ([xshift=0.45*\hnode, anchor=west]y0.east) {$y_1$};
				\node[ynode] (y2) at ([xshift=0.45*\hnode]y1.east) {$y_2$};
				\node[ynode] (y3) at ([xshift=0.45*\hnode]y2.east) {$y_3$};
				
				\begin{pgfonlayer}{background}
				\path[fill=red!20] (a00.south west) rectangle (a00.north east);
				\path[fill=red!20] (a10.south west) rectangle (a10.north east);
				\path[fill=red!20] (a20.south west) rectangle (a20.north east);
				\path[fill=red!20] (a30.south west) rectangle (a30.north east);
				\end{pgfonlayer}	
				
				\end{scope}
				\end{tikzpicture}
			}
			&
			
			\subfloat [%\small{Our}
			] {
				\begin{tikzpicture}
				\begin{scope}	
				
				\setlength{\hnode}{1.2em}
				\tikzstyle{elementnode} = [rectangle,anchor=center]
				\tikzstyle{ynode} = [font=\small]
				\tikzstyle{xnode} = [left,font=\small,anchor=east]
				
				% weight
				\foreach \i / \j / \c in
				{0/3/1.8,
					0/2/.4,   1/2/1.2,
					0/1/.3,  1/1/.2, 2/1/.8,
					0/0/.1,   1/0/.3,  2/0/.5, 3/0/.7}
				\node[elementnode,minimum size=1*\hnode,inner sep=0.1pt, draw] (a\i\j) at (\hnode*\i, \hnode*\j) {\scriptsize{\c}};
				
				% x
				\node[xnode] (x1) at (a03.west) {$x_1$};
				\node[xnode] (x2) at (a02.west) {$x_2$};
				\node[xnode] (x3) at (a01.west) {$x_3$};
				\node[xnode] (x4) at (a00.west) {$x_4$};
				
				% y
				\node[ynode,anchor=south] (y0) at (a03.north) {$y_0$};
				\node[ynode] (y1) at ([xshift=0.45*\hnode, anchor=west]y0.east) {$y_1$};
				\node[ynode] (y2) at ([xshift=0.45*\hnode]y1.east) {$y_2$};
				\node[ynode] (y3) at ([xshift=0.45*\hnode]y2.east) {$y_3$};
				
				\begin{pgfonlayer}{background}
				\path[fill=red!20] (a00.south west) rectangle (a00.north east);
				\path[fill=red!20] (a10.south west) rectangle (a10.north east);
				\path[fill=red!20] (a20.south west) rectangle (a20.north east);
				\path[fill=red!20] (a30.south west) rectangle (a30.north east);
				\path[fill=red!20] (a01.south west) rectangle (a01.north east);
				\path[fill=red!20] (a11.south west) rectangle (a11.north east);
				\path[fill=red!20] (a21.south west) rectangle (a21.north east);
				\path[fill=red!20] (a02.south west) rectangle (a02.north east);
				\path[fill=red!20] (a12.south west) rectangle (a12.north east);
				\path[fill=red!20] (a03.south west) rectangle (a03.north east);
				
				\end{pgfonlayer}
				
				\end{scope}
				\end{tikzpicture}
			}
			\\
			
		\end{tabular}
	\end{center}
	
	\begin{center}
		\vspace{-.5em}
		\caption{Connection weights for 3-layer encoder: (a) residual connection \cite{he2016deep}, (b) dense residual connection \cite{britz2017massive,dou2018exploiting}, (c) multi-layer representation fusion \cite{wang2018multi}/transparent attention \cite{bapna2018training} and (d) our approach. $y_0$ denotes the input embedding. Red denotes the weights are learned by model.}
		\label{fig:diff}
		\vspace{-1.2em}
	\end{center}
\end{figure*}
%% FIGURE ENDS
%%%%%%%%%%%%%%%%

The residual network is the most common approach to learning deep networks, and plays an important role in Transformer. In principle, residual networks can be seen as instances of the ordinary differential equation (ODE), behaving like the forward Euler discretization with an initial value  \cite{chang2018multilevel,NIPS2018_7892}.
Euler's method is probably the most popular first-order solution to ODE. But it is not yet accurate enough. A possible reason is that only one previous step is used to predict the current value \footnote{Some of the other single-step methods, e.g. the Runge-Kutta method, can obtain a higher order by taking several intermediate steps \cite{Butcher_2003}. Higher order generally means more accurate.}\cite{Butcher_2003}.
In MT, the single-step property of the residual network makes the model ``forget'' distant layers \cite{wang2018multi}. As a result, there is no easy access to features extracted from lower-level layers if the model is very deep.

Here, we describe a model which makes direct links with all previous layers and offers efficient access to lower-level representations in a deep stack. We call it dynamic linear combination of layers (\name). The design is inspired by the linear multi-step method (LMM) in numerical ODE \cite{ascher1998computer}. Unlike Euler's method, LMM can effectively reuse the information in the previous steps by linear combination to achieve a higher order.
Let $\{y_0,...,y_l\}$ be the output of layers $0 \sim l$. The input of layer $l+1$ is defined to be
\begin{equation}
\label{eq-res-general-extend}
%y_l &= h(x_l) + \mathcal{F}(x_l; \theta_l), \\
x_{l+1} = \mathcal{G}(y_0, \ldots, y_l)
\end{equation}

\noindent where $\mathcal{G}(\cdot)$ is a linear function that merges previously generated values $\{y_0,...,y_l\}$ into a new value. For pre-norm Transformer, we define $\mathcal{G}(\cdot)$ to be
\begin{equation}
\label{eq-dla}
%\begin{split}
\mathcal{G}(y_0, \ldots, y_l) = \sum_{k=0}^{l} W^{(l+1)}_k\textrm{LN}(y_k)
%\end{split}
\end{equation}
\noindent where $W_k^{l+1} \in \mathbb{R}$ is a learnable scalar and weights each incoming layer in a linear manner. Eq. (\ref{eq-dla}) provides a way to learn preference of layers in different levels of the stack. Even for the same incoming layer, its contribution to succeeding layers could be different (e.g. $W_k^i \ne W_k^k$)
%\footnote{There is no requirement of normalization here, e.g. $\sum_{k=0}^{k=l-1}W^l_k=1$ }
. Also, the method is applicable to the post-norm Transformer model. For post-norm, $\mathcal{G}(\cdot)$ can be redefined as:
\begin{equation}
\label{eq-dla-post}
\mathcal{G}(y_0, \ldots, y_l) = \textrm{LN}\big(\sum_{k=0}^{l} W^{(l+1)}_k y_k\big)
\end{equation}

\begin{itemize}[leftmargin=*, label=]
	\item \textbf{Comparison to LMM. }
	\name differs from LMM in two aspects, though their fundamental model is the same. First, \name learns weights in an end-to-end fashion rather than assigning their values deterministically, e.g. by polynomial interpolation. This offers a more flexible way to control the model behavior. Second, \name has an arbitrary size of the past history window, while LMM generally takes a limited history into account \cite{loczi2018exact}.
	Also, recent work shows successful applications of LMM in computer vision, but only two previous steps are used in their LMM-like system \cite{lu18d}.
	
	\item \textbf{Comparison to existing neural methods. }
	Note that \name is a very general approach. For example, the standard residual network is a special case of \name, where $W_l^{l+1} = 1$, and $W_k^{l+1}=0$ for $k < l$. Figure (\ref{fig:diff}) compares different methods of connecting a 3-layer network. We see that the densely residual network is a fully-connected network with a uniform weighting schema \cite{britz2017massive,dou2018exploiting}. Multi-layer representation fusion \cite{wang2018multi} and transparent attention (call it TA) \cite{bapna2018training} methods can learn a weighted model to fuse layers but they are applied to the topmost layer only. The \name model can cover all these methods. It provides ways of weighting and connecting layers in the entire stack. We emphasize that although the idea of weighting the encoder layers by a learnable scalar is similar to TA, there are two key differences: 1) Our method encourages earlier interactions between layers during the encoding process, while the encoder layers in TA are combined until the standard encoding process is over; 2) For an encoder layer, instead of learning a unique weight for each decoder layer like TA, we make a separate weight for each successive encoder layers. In this way, we can create more connections between layers\footnote{Let the encoder depth be $M$ and the decoder depth be $N$ ($M>N$ for a deep encoder model). Then TA newly adds $\mathcal{O}(M \times N)$ connections, which are fewer than ours of $\mathcal{O}(M^2)$}.
\end{itemize}

\section{Experimental Setup}

% cancel SacreBLEU, add Epochs
\begin{table*}[thb]
	%\small
	\begin{center}
		\renewcommand\arraystretch{1}
		\begin{tabular}{l|l | c  c  c  c | c c}
			\toprule[1pt]
			
			%% German-English Translation Result
			\multicolumn{2}{c ||}{\textbf{Model}} & \multicolumn{1}{c| }{\textbf{Param.}} &
			\multicolumn{1}{c }{\textbf{Batch}} & \multicolumn{1}{c }{\textbf{Updates}} &
			\multicolumn{1}{c |}{\textbf{$^\dagger$Times}} &
			\multicolumn{1}{c }{\textbf{BLEU}} &
			\multicolumn{1}{c }{\textbf{$\Delta$}} \\
			
			\multicolumn{2}{c ||}{} & \multicolumn{1}{c| }{} &
			\multicolumn{1}{c }{($\times$4096)} & \multicolumn{1}{c }{($\times$100k)} &
			\multicolumn{1}{c |}{} &
			\multicolumn{1}{c }{} &
			\multicolumn{1}{c }{} \\
			
			\hline %\hline
			
			\multicolumn{2}{c||}{\footnotesize{\newcite{vaswani2017attention} (Base)}} 		
			& \multicolumn{1}{r|}{\hspace{0.1cm}65M$^{}$} & 1 & 1 &  \footnotesize{reference} & 27.3 & - \\
			
			\multicolumn{2}{c||}{\footnotesize{\newcite{bapna2018training}-deep (Base, 16L)}} 		
			& \multicolumn{1}{r|}{137M} & - & -  & - & 28.0 & - \\
			\cdashline{1-8}
			
			\multicolumn{2}{c||}{\footnotesize{\newcite{vaswani2017attention} (Big)}} 		
			& \multicolumn{1}{r|}{213M} & 1 & 3 &  3x & 28.4 & - \\
			
			\multicolumn{2}{c||}{\footnotesize{\citet{chen2018best} (Big)}} 		
			& \multicolumn{1}{r|}{379M} & 16 & $^\dagger$0.075 &  1.2x & 28.5 & - \\
			
			\multicolumn{2}{c||}{\footnotesize{\citet{he2018layer} (Big)}} 		
			& \multicolumn{1}{r|}{$^\dagger$210M} & 1 & - & - & 29.0 & - \\
			
			\multicolumn{2}{c||}{\footnotesize{\citet{shaw2018self} (Big)}} 		
			& \multicolumn{1}{r|}{$^\dagger$210M} & 1 & 3 & 3x & 29.2 & - \\
			
			\multicolumn{2}{c||}{\footnotesize{\citet{dou2018exploiting} (Big)}} 		
			& \multicolumn{1}{r|}{356M} & 1 & - & - & 29.2 & - \\
			
			\multicolumn{2}{c||}{\footnotesize{\newcite{DBLP:conf/wmt/OttEGA18} (Big)}} 		
			& \multicolumn{1}{r|}{210M} & 14 & 0.25 & 3.5x & 29.3 & - \\
			\hline
			
			\multicolumn{1}{l|}{\multirow{5}{*}{post-norm}} &
			\multicolumn{1}{l||}{\footnotesize{Transformer (Base)}}
			& \multicolumn{1}{r|}{\hspace{0.1cm}62M} & 1 & 1 & 1x & 27.5 & \footnotesize{reference} \\
			~ & \multicolumn{1}{l||}{\footnotesize{Transformer (Big)}}
			& \multicolumn{1}{r|}{211M} & 1 & 3 & 3x & 28.8 & +1.3  \\
			~ & \multicolumn{1}{l||}{\footnotesize{Transformer-deep (Base, 20L)}}
			& \multicolumn{1}{r|}{106M} & 2 & 0.5 & 1x &  \footnotesize{failed} & \footnotesize{failed}  \\
			\cdashline{2-8}
			~ & \multicolumn{1}{l||}{\footnotesize{DLCL (Base)}}
			& \multicolumn{1}{r|}{\hspace{0.1cm}62M} & 1 & 1 & 1x & 27.6 & +0.1  \\
			~ & \multicolumn{1}{l||}{\footnotesize{DLCL-deep (Base, 25L)}}
			& \multicolumn{1}{r|}{121M} & 2 & 0.5 & 1x & 29.2 & +1.7 \\
			\hline
			
			\multicolumn{1}{l|}{\multirow{5}{*}{pre-norm}} & \multicolumn{1}{l||}{\footnotesize{Transformer (Base)}}
			& \multicolumn{1}{r|}{\hspace{0.1cm}62M} & 1 & 1 & 1x & 27.1 & \footnotesize{reference} \\
			~ & \multicolumn{1}{l||}{\footnotesize{Transformer (Big)}}
			& \multicolumn{1}{r|}{211M} & 1 & 3 &  3x &  28.7  & +1.6 \\
			~ & \multicolumn{1}{l||}{\footnotesize{Transformer-deep (Base, 20L)}}
			& \multicolumn{1}{r|}{106M} & 2 & 0.5 &  1x & 28.9 & +1.8 \\
			\cdashline{2-8}
			~ & \multicolumn{1}{l||}{\footnotesize{DLCL (Base)}}
			& \multicolumn{1}{r|}{\hspace{0.1cm}62M} & 1 & 1 &  1x & 27.3 & +0.2  \\
			~ & \multicolumn{1}{l||}{\footnotesize{DLCL-deep (Base, 30L)}}
			& \multicolumn{1}{r|}{137M} & 2 & 0.5 &  1x & \textbf{29.3} & \textbf{+2.2} \\
			
			\bottomrule[1.pt]
		\end{tabular}
		
		\vspace{-0.0em}
		\caption{BLEU scores [\%] on English-German translation. \texttt{Batch} indicates the corresponding batch size if running on 8 GPUs. \texttt{Times} $\propto$ \texttt{Batch}$\times$\texttt{Updates}, which can be used to approximately measure the required training time. $^\dagger$ denotes an estimate value. Note that ``-deep'' represents the best-achieved result as depth changes.}
		\label{table:de2en-result}
		\vspace{-0.5em}
	\end{center}
\end{table*}

\noindent We first evaluated our approach on WMT'16 English-German (En-De) and NIST'12 Chinese-English (Zh-En-Small) benchmarks respectively. To make the results more convincing, we also experimented on a larger WMT'18 Chinese-English  dataset (Zh-En-Large) with data augmentation by back-translation \cite{sennrich2016improving}.

\subsection{Datasets and Evaluation}
\noindent For the En-De task, to compare with \citet{vaswani2017attention}'s work, we use the same 4.5M pre-processed data \footnote{\url{https://drive.google.com/uc?export=download&id=0B_bZck-ksdkpM25jRUN2X2UxMm8}}, which has been tokenized and jointly byte pair encoded (BPE) \cite{sennrich2015neural} with 32k merge operations using a shared vocabulary \footnote{The tokens with frequencies less than 5 are filtered out from the shared vocabulary.}. We use \textit{newstest2013} for validation and \textit{newstest2014} for test.

For the Zh-En-Small task, we use parts of the bitext provided within NIST'12 OpenMT\footnote{LDC2000T46, LDC2000T47, LDC2000T50, LDC2003E14, LDC2005T10, LDC2002E18, LDC2007T09, LDC2004T08}. We choose NIST MT06 as the validation set, and MT04, MT05, MT08 as the test sets. All the sentences are word segmented by the tool provided within NiuTrans \cite{xiao2012niutrans}. We remove the sentences longer than 100 and end up with about 1.9M sentence pairs. Then BPE with 32k operations is used for both sides independently, resulting in a 44k Chinese vocabulary and a 33k English vocabulary respectively.

For the Zh-En-Large task, we use exactly the same 16.5M dataset as \citet{wang2018niutrans}, composing of 7.2M-sentence CWMT corpus, 4.2M-sentence UN and News-Commentary combined corpus, and back-translation of 5M-sentence monolingual data from NewsCraw2017. We refer the reader to \citet{wang2018niutrans} for the details.

For evaluation, we first average the last 5 checkpoints, each of which is saved at the end of an epoch. And then we use beam search with a beam size of 4/6 and length penalty of 0.6/1.0 for En-De/Zh-En tasks respectively. We measure case-sensitive/insensitive tokenized BLEU by \textit{multi-bleu.perl} for En-De and Zh-En-Small respectively, while case-sensitive detokenized BLEU is reported by the official evaluation script \textit{mteval-v13a.pl} for Zh-En-Large. Unless noted otherwise we run each experiment three times with different random seeds and report the mean of the BLEU scores across runs\footnote{Due to resource constraints, all experiments on Zh-En-Large task only run once.}.

\subsection{Model and Hyperparameters}

\begin{table}[t]
	\begin{center}
		%\small
		\renewcommand\arraystretch{1.}
		\begin{tabular}{l | l | c }
			%\hline
			%\hline
			\toprule[1pt]
			
			\multicolumn{2}{c }{\textbf{Model (Base, 16L)}} &
			\multicolumn{1}{c }{\textbf{BLEU}} \\
			\hline %\hline
			
			\multicolumn{1}{l|}{\multirow{3}{*}{post-norm}} &
			\multicolumn{1}{l }{\newcite{bapna2018training}} & 28.0 \\
			~ &  \multicolumn{1}{l }{Transformer} & failed \\
			~ &  \multicolumn{1}{l }{DLCL} & 28.4 \\
			\cdashline{1-3}	
			\multicolumn{1}{l|}{\multirow{2}{*}{pre-norm}} &
			\multicolumn{1}{l }{Transformer} & 28.0 \\
			~ & \multicolumn{1}{l}{DLCL} & 28.2	\\
			\bottomrule[1pt]
			
		\end{tabular}
		
		\vspace{-0.5em}
		\caption{Compare with \citet{bapna2018training} on WMT'16 English-German translation under a 16-layer encoder.}
		\label{table:16l-en2de-result}
		\vspace{-1.0em}
	\end{center}
\end{table}

% exclude MT12 result
\begin{table*}[thb]
	\begin{center}
		%\small
		\renewcommand\arraystretch{1.}
		\begin{tabular}{l | c | c  c  c  c c}
			%\hline
			%\hline
			\toprule[1pt]
			
			%% German-English Translation Result
			%	\multicolumn{6}{l}{\textbf{NIST12 Chinese$\to$English}} \\ \hline
			\multicolumn{1}{c ||}{\textbf{Model (pre-norm)}} &
			\multicolumn{1}{c |}{\textbf{Param.}} &
			\multicolumn{1}{c }{\textbf{Valid.}} &
			\multicolumn{1}{c }{\textbf{MT04}} &
			\multicolumn{1}{c }{\textbf{MT05}}&
			\multicolumn{1}{c }{\textbf{MT08}}&
			\multicolumn{1}{c }{\textbf{Average}} \\
			\hline %\hline
			
			\multicolumn{1}{l||}{Transformer (Base)} 		
			& \multicolumn{1}{r|}{\hspace{0.1cm}84M} 	& 51.27 	& 54.41		& 49.43	& 45.33	&  49.72 \\
			\multicolumn{1}{l||}{Transformer (Big)}  	
			& \multicolumn{1}{r|}{257M} 	& 52.30 & 55.37 	& 52.21	& 47.40	&  51.66 \\
			\multicolumn{1}{l||}{Transformer-deep (Base, 25L)}  	
			& \multicolumn{1}{r|}{144M} 	& 52.50	& 55.80 	& 51.98	& 47.26	&  51.68\\
			%\hline
			\cdashline{1-7}	
			
			\multicolumn{1}{l||}{DLCL (Base)}
			& \multicolumn{1}{r|}{\hspace{0.1cm}84M} 	& 51.61 & 54.91 	& 50.58 & 46.11	&  50.53\\
			\multicolumn{1}{l||}{DLCL-deep (Base, 25L)}
			& \multicolumn{1}{r|}{144M} 	& \textbf{53.57} &  \textbf{55.91} & \textbf{52.30} & \textbf{48.12} &  \textbf{52.11} \\
			
			\bottomrule[1pt]
			
		\end{tabular}
		
		\vspace{-0.5em}
		\caption{BLEU scores [\%] on NIST'12 Chinese-English translation.}
		\label{table:zh2en-result}
		\vspace{-1.0em}
	\end{center}
\end{table*}

\noindent All experiments run on \emph{fairseq-py}\footnote{\url{https://github.com/pytorch/fairseq}} with 8 NVIDIA Titan V GPUs. For the post-norm Transformer baseline, we replicate the model setup of \citet{vaswani2017attention}. All models are optimized by Adam \cite{kingma2014adam} with  \textrm{$\beta_1$} = 0.9, \textrm{$\beta_2$} = 0.98, and \textrm{$\epsilon$} = 10$^{-8}$. In training warmup (\texttt{warmup} = 4000 steps), the learning rate linearly increases from 10$^{-7}$ to \texttt{lr} =7$\times$10$^{-4}$/5$\times$10$^{-4}$ for Transformer-Base/Big respectively, after which it is decayed proportionally to the inverse square root of the current step. Label smoothing \textrm{$\varepsilon_{ls}$}=0.1 is used as regularization.

For the pre-norm Transformer baseline, we follow the setting as suggested in tensor2tensor\footnote{\url{https://github.com/tensorflow/tensor2tensor}}. More specifically, the attention dropout P$_{att}$ = 0.1 and feed-forward dropout P$_{ff}$ = 0.1 are additionally added. And some hyper-parameters for optimization are changed accordingly: \textrm{$\beta_2$} = 0.997, \texttt{warmup} = 8000 and \texttt{lr} = 10$^{-3}$/7$\times$10$^{-4}$ for Transformer-Base/Big respectively.

For both the post-norm and pre-norm baselines, we batch sentence pairs by approximate length and restrict input and output tokens per batch to \texttt{batch} = 4096 per GPU. We set the update steps according to corresponding data sizes. More specifically, the Transformer-Base/Big is updated for 100k/300k steps on the En-De task as \citet{vaswani2017attention}, 50k/100k steps on the Zh-En-Small task, and 200k/500k steps on the Zh-En-Large task.

In our model, we use the dynamic linear combination of layers for both encoder and decoder. For efficient computation, we only combine the output of a complete layer rather than a sub-layer. It should be noted that for deep models (e.g. $L$ $\ge$ 20), it is hard to handle a full batch in a single GPU due to memory size limitation. We solve this issue by accumulating
gradients from two small batches (e.g. batch = 2048) before each update \cite{DBLP:conf/wmt/OttEGA18}. In our primitive experiments, we observed that training with  larger batches and learning rates worked well for deep models. Therefore all the results of deep models are reported with \texttt{batch} = 8192, \texttt{lr} = 2$\times$10$^{-3}$ and \texttt{warmup} = 16,000 unless otherwise stated. For fairness, we only use half of the updates of baseline (e.g. \texttt{update} = 50k) to ensure the same amount of data that we actually see in training. We report the details in Appendix B.

\section{Results}
\label{sec:results}

\subsection{Results on the En-De Task}
\label{sec:en2de_result}

\begin{table*}[thb]
	\begin{center}
		%\small
		\renewcommand\arraystretch{1.}
		\begin{tabular}{l | c | c c c}
			%\hline
			%\hline
			\toprule[1pt]
			
			%% German-English Translation Result
			%	\multicolumn{6}{l}{\textbf{NIST12 Chinese$\to$English}} \\ \hline
			\multicolumn{1}{c ||}{\textbf{Model}} &
			\multicolumn{1}{c |}{\textbf{Param.}} &
			\multicolumn{1}{c }{\textbf{newstest17}} &
			\multicolumn{1}{c }{\textbf{newstest18}} &
			\multicolumn{1}{c }{$\Delta_{avg.}$} \\
			\hline %\hline
			
			%\multicolumn{1}{l}{\multirow{3}{*}{Pre-normalize Baselines}} &
			%\multicolumn{4}{c }{Existing Systems} \\
			%\hline
			\multicolumn{1}{l||}{\citet{wang2018niutrans} (post-norm, Base)} 		
			& \multicolumn{1}{r|}{\hspace{0.1cm}102.1M} 	& 25.9 	& - & - \\
			\cdashline{1-5}	
			
			\multicolumn{1}{l||}{pre-norm Transformer (Base)} 		
			& \multicolumn{1}{r|}{\hspace{0.1cm}102.1M} 	& 25.8 	& 25.9 & \footnotesize{reference}\\
			\multicolumn{1}{l||}{pre-norm Transformer (Big)}  	
			& \multicolumn{1}{r|}{292.4M} 	& 26.4	&  27.0 & +0.9 \\
			\multicolumn{1}{l||}{pre-norm DLCL-deep (Base, 25L)}  	
			& \multicolumn{1}{r|}{161.5M} 	& 26.7	& 27.1 & +1.0 \\
			\multicolumn{1}{l||}{pre-norm DLCL-deep (Base, 30L)}
			& \multicolumn{1}{r|}{177.2M} 	& \textbf{26.9} &  \textbf{27.4} & \textbf{+1.3} \\
			\bottomrule[1pt]
			
		\end{tabular}
		
		\vspace{-0.5em}
		\caption{BLEU scores [\%] on  WMT'18 Chinese-English translation.}
		\label{table:large-zh2en-result}
		\vspace{-1.0em}
	\end{center}
\end{table*}

\noindent In Table~\ref{table:de2en-result}, we first report results on WMT En-De where we compare to the existing systems based on self-attention. Obviously, while almost all previous results based on Transformer-Big (marked by Big) have higher BLEU than those based on Transformer-Base (marked by Base), larger parameter size and longer training epochs are required.

As for our approach, considering the post-norm case first, we can see that our Transformer baselines are superior to \citet{vaswani2017attention} in both \texttt{Base} and \texttt{Big} cases. When increasing the encoder depth, e.g. $L$ = 20, the vanilla Transformer failed to train, which is consistent with \citet{bapna2018training}. We attribute it to the vanishing gradient problem based on the observation that the gradient norm in the low layers (e.g. embedding layer) approaches 0. On the contrary, post-norm \name solves this issue and achieves the best result when $L$ = 25.

The situation changes when switching to pre-norm. While it slightly underperforms the post-norm counterpart in shallow networks, pre-norm Transformer benefits more from the increase in encoder depth. More concretely, pre-norm Transformer achieves optimal result when $L$=20 (see Figure~\ref{fig:enc-depth}(a)), outperforming the 6-layer baseline by 1.8 BLEU points. It indicates that pre-norm is easier to optimize than post-norm in deep networks.
Beyond that, we successfully train a 30-layer encoder by our method, resulting in a further improvement of 0.4 BLEU points. This is 0.6 BLEU points higher than the pre-norm Transformer-Big. It should be noted that although our best score of 29.3 is the same as \citet{DBLP:conf/wmt/OttEGA18}, our approach only requires 3.5X fewer training epochs than theirs.

To fairly compare with \textit{transparent attention} (TA) \cite{bapna2018training}, we separately list the results using a 16-layer encoder in Table~\ref{table:16l-en2de-result}. It can be seen that pre-norm Transformer obtains the same BLEU score as TA without the requirement of complicated attention design. However, \name in both post-norm and pre-norm cases outperform TA. It should be worth that TA achieves the best result when encoder depth is 16, while we can further improve performance by training deeper encoders.

\subsection{Results on the Zh-En-Small Task}
\label{sec:zh2en_result}

\noindent Seen from the En-De task, pre-norm is more effective than the post-norm counterpart in deep networks. Therefore we evaluate our method in the case of pre-norm on the Zh-En task. As shown in Table~\ref{table:zh2en-result}, firstly \name is superior to the baseline when the network's depth is shallow. Interestingly, both Transformer and \name achieve the best results when we use a 25-layer encoder. The 25-layer Transformer can approach the performance of Transformer-Big, while our deep model outperforms it by about 0.5 BLEU points under the equivalent parameter size. It confirms that our approach is a good alternative to Transformer no matter how deep it is.

\subsection{Results on the Zh-En-Large Task}
\label{sec:large_zh2en_result}

\begin{figure}[tbp]
	\begin{center}
		\renewcommand\arraystretch{0.0}
		\setlength{\tabcolsep}{1pt}
		\begin{tabular}{cc}
			
			%\hline
			\multicolumn{2}{c}
			{
				\begin{tikzpicture}
				\scriptsize
				\node (l0) at (0,0) {};
				\draw[ugreen, dashed] (0,0) -- plot[](0.25,0) -- (0.5,0) node[right] (l1) {Base-6L};
				\draw[lyellow, dashed] (2,0) -- plot[](2.25,0) -- (2.5,0) node[right] (l2) {Big-6L};
				\draw[red] (4,0) -- plot[mark=otimes*](4.25,0) -- (4.5,0) node[right] (l3) {Transformer};
				\draw[blue] (6.2,0) -- plot[mark=square](6.45,0) -- (6.7,0) node[right] (l4) {\plainname};
				\begin{pgfonlayer}{background}
				\node[rectangle,draw,inner sep=1pt] [fit = (l0) (l1) (l2) (l3) (l4)] {};
				\end{pgfonlayer}	
				\end{tikzpicture}
			} \\
			%\hline
			
			\subfloat[\footnotesize{WMT En-De}]
			{
				\begin{tikzpicture}{baseline}
				\scriptsize{
					\begin{axis}[
					ylabel near ticks,
					width=.24\textwidth,
					height=.2\textwidth,
					legend style={at={(0.62,0.33)}, anchor=south west},
					xlabel={},
					ylabel={\scriptsize{BLEU Score}},
					ylabel style={yshift=-0em},xlabel style={yshift=0.0em},
					yticklabel style={/pgf/number format/precision=1,/pgf/number format/fixed zerofill},
					ymin=26.50,ymax=29.50, ytick={26.5, 27.0, 27.5, 28.0, 28.5, 29.0, 29.5},
					xmin=6,xmax=35,xtick={6,16,20,25,30,35},
					legend style={yshift=-12pt, legend plot pos=left,font=\tiny,cells={anchor=west}}
					]
					
					\addplot[red,mark=otimes*,line width=0.5pt] coordinates {(6,27.10) (16,28.00) (20,28.90) (25,28.78) (30,28.67) (35,28.30)};
					%\addlegendentry{pre-norm}
					
					\addplot[blue,mark=square,line width=0.5pt] coordinates {(6,27.30) (16,28.20) (20,28.80) (25,29.1) (30,29.30) (35,28.85)};
					%\addlegendentry{\plainname}
					
					% valid. blue in baseline
					\addplot[dashed,domain=6:35, ugreen, line width=0.5pt]{27.1};
					%\addlegendentry{base pre.}
					
					% test blue in baseline
					\addplot[dashed,domain=6:35, lyellow, line width=0.5pt]{28.7};
					%\addlegendentry{big pre.}
					
					\end{axis}
				}
				\label{fig:enc_depth_en2de}
				\end{tikzpicture}
			}
			&
			\subfloat [\footnotesize{NIST Zh-En}]
			{
				\begin{tikzpicture}{baseline}
				\scriptsize{
					\begin{axis}[
					ylabel near ticks,
					width=.24\textwidth,
					height=.2\textwidth,
					legend style={at={(0.62,0.3)}, anchor=south west},
					xlabel={},
					ylabel={\scriptsize{BLEU Score}},
					ylabel style={yshift=-0em},xlabel style={yshift=0.0em},
					yticklabel style={/pgf/number format/precision=1,/pgf/number format/fixed zerofill},
					ymin=49.5,ymax=52.5,ytick={49.0, 49.5, 50.0, 50.5, 51.0, 51.5, 52.0, 52.5},
					xmin=6,xmax=30,xtick={6,16,20,25,30},
					legend style={yshift=-12pt, legend plot pos=left,font=\tiny,cells={anchor=west}}
					]
					
					\addplot[red,mark=otimes*,line width=0.5pt] coordinates {(6,49.7) (16,51.37) (20,51.66) (25,51.68) (30,51.7)};
					%\addlegendentry{pre-norm}
					\addplot[blue,mark=square,line width=0.5pt] coordinates {(6,50.53) (16,51.45) (20,51.68) (25,52.11) (30,52.04)};
					%\addlegendentry{\plainname}
					
					% valid. blue in baseline
					\addplot[dashed,domain=6:30, ugreen, line width=0.5pt]{49.72};
					
					% test blue in baseline
					\addplot[dashed,domain=6:30, lyellow, line width=0.5pt]{51.66};
					
					\end{axis}
				}
				\label{fig:enc_depth_zh2en}
				\end{tikzpicture}
			}
			
		\end{tabular}
	\end{center}
	
	\begin{center}
		\vspace{-0.5em}
		\caption{ BLEU scores [\%] against the encoder depth for pre-norm Transformer and pre-norm DLCL on English-German and Chinese-English tasks. }
		\label{fig:enc-depth}
		\vspace{-1.0em}
	\end{center}
\end{figure}

\noindent While deep Transformer models, in particular the deep pre-norm DLCL, show better results than Transformer-Big on En-De and Zh-En-Small tasks, both data sets are relatively small, and the improved performance over Transformer-Big might be partially due to over-fitting in the wider model. For a more challenging task , we report the results on Zh-En-Large task in Table~\ref{table:large-zh2en-result}. We can see that the 25-layer pre-norm DLCL slightly surpassed Transformer-Big, and the superiority is bigger when using a 30-layer encoder. This result indicates that the claiming of the deep network defeating Transformer-Big is established and is not affected by the size of the data set.

\section{Analysis}

\subsection{Effect of Encoder Depth}

\begin{figure}[t]
	\begin{center}
		\setlength{\tabcolsep}{1pt}
		\begin{tikzpicture}{baseline}
		\footnotesize{
			\begin{axis}[
			width=.4\textwidth,
			height=.2\textwidth,
			legend style={at={(0.1,1.3)}, anchor=south west},
			xlabel={},
			ylabel={\footnotesize{Speed}},
			ylabel style={yshift=-0em},xlabel style={yshift=0.0em},
			yticklabel style={/pgf/number format/precision=0,/pgf/number format/fixed zerofill},
			ymin=1800,ymax=2700, ytick={1800, 2000, 2200, 2400, 2600, 2800},
			xmin=6,xmax=30,xtick={6,16,20,25,30},
			legend columns=3,
			legend style={xshift=-3pt, yshift=-12pt, legend plot pos=left,font=\tiny,cells={anchor=west},
				% the /tikz/ prefix is necessary here...
				% otherwise, it might end-up with `/pgfplots/column 2`
				% which is not what we want. compare pgfmanual.pdf
				/tikz/column 6/.style={column sep=7pt,}}
			]
			
			% 6 layer of base model
			\addplot[dashed,domain=0.5:35, ugreen, line width=0.8pt]{2589.2};
			\addlegendentry{Base-6L}
			
			% 6 layer of big model
			\addplot[dashed,domain=0.5:35, lyellow, line width=0.8pt]{1896};
			\addlegendentry{Big-6L}
			
			%\addplot[red,mark=otimes*,line width=0.5pt] coordinates {(6,26.90) (16,27.80) (20,28.70) (25,28.58) (30,28.47) (35,28.10)};
			%\addlegendentry{pre-norm}
			\addplot[blue,mark=square,line width=0.8pt] coordinates {(6,2502.2) (16,2303) (20,2237.37) (25,2186) (30,2109)};
			\addlegendentry{DLCL}
			
			\end{axis}
		}
		\end{tikzpicture}
	\end{center}
	
	\begin{center}
		\vspace{-0.5em}
		\caption{GPU generation speed (target tokens/sec.) against the depth of encoder for pre-norm DLCL on English-German task (batch size = 32, beam size = 4).}
		\label{fig:enc-depth-speed}
		\vspace{-1.em}
	\end{center}
\end{figure}

\noindent In Figure~\ref{fig:enc-depth}, we plot BLEU score as a function of encoder depth for pre-norm Transformer and \name on En-De and Zh-En-Small tasks. First of all, both methods benefit from an increase in encoder depth at the beginning. Remarkably, when the encoder depth reaches 20, both of the two deep models can achieve comparable performance to Transformer-Big, and even exceed it when the encoder depth is further increased in \name.
Note that pre-norm Transformer degenerates earlier and is less robust than \name when the depth is beyond 20. However, a deeper network ($>$30 layers) does not bring more benefits. Worse still, deeper networks consume a lot of memory, making it impossible to train efficiently.

We also report the inference speed on GPU in Figure~\ref{fig:enc-depth-speed}. As expected, the speed decreases linearly with the number of encoder layers. Nevertheless, our system with a 30-layer encoder is still faster than Transformer-Big, because the encoding process is independent of beam size, and runs only once. In contrast, the decoder suffers from severe autoregressive problems.

\subsection{Effect of Decoder Depth}
\begin{table}[thb]
	%\small
	\begin{center}
		\renewcommand\arraystretch{1.}
		\begin{tabular}{c c | c c }
			%\hline
			%\hline
			\toprule[1pt]
			
			\multicolumn{1}{c }{\footnotesize{\textbf{Enc. Depth}} } &
			\multicolumn{1}{c |}{\footnotesize{\textbf{Dec. Depth}}} &
			\multicolumn{1}{c }{\footnotesize{\textbf{BLEU}}} &
			\multicolumn{1}{c }{\footnotesize{\textbf{Speed}} } \\
			\hline %\hline
			
			\multicolumn{1}{c }{6} 	& 4 & 27.12 & {3088.3} \\
			\multicolumn{1}{c }{6}  & 6 & 27.33 & {2589.2} \\
			\multicolumn{1}{c }{6}  & 8 & 27.42 & {2109.6} \\
			
			\bottomrule[1pt]
		\end{tabular}
		
		\vspace{-0.5em}
		\caption{Tokenized BLEU scores [\%] and GPU generation speed (target tokens per second)  in pre-norm Transformer (Base) on the test set of WMT English-German (batch size = 32, beam size = 4).}
		\label{table:dec-depth}
		\vspace{-1.em}
	\end{center}
\end{table}

%期望的结果是表明加深，性能增加不明显，且会显著降低解码效率
\noindent Table~\ref{table:dec-depth} shows the effects of decoder depth on BLEU and inference speed on GPU. Different from encoder, increasing the depth of decoder only yields a slight BLEU improvement, but the cost is high: for every two layers added, the translation speed drops by approximate 500 tokens evenly. It indicates that exploring deep encoders may be more promising than deep decoders for NMT.

\subsection{Ablation Study}
\label{sec:ablation}

\begin{table}[t]
	%\small
	\begin{center}
		\begin{tabular}{c c }
			%\hline
			%\hline
			\toprule[1pt]
			
			\multicolumn{1}{c }{\footnotesize{\textbf{Model}} } &
			\multicolumn{1}{c }{\footnotesize{\textbf{BLEU}}} \\
			\hline %\hline
			
			\multicolumn{1}{l }{pre-norm DLCL-20L} & 28.80  \\
			\multicolumn{1}{l }{\hspace{0.1cm}- layer norm.}  & 28.67 \\
			\multicolumn{1}{l }{\hspace{0.1cm}- learnable weight (fix 1)}  & 28.22 \\
			\multicolumn{1}{l }{\hspace{0.1cm}- learnable weight (fix 1/N)}  & 28.51 \\
			\multicolumn{1}{l }{\hspace{0.3cm}- layer norm.}  & 28.23 \\
			\hline %\hline
			
			\multicolumn{1}{l }{pre-norm Transformer-Base} & 27.11 \\
			\multicolumn{1}{l }{\hspace{0.1cm}+ big encoder} & 27.59 \\
			\hline %\hline
			
			\multicolumn{1}{l }{pre-norm Transformer-Big} & 28.72 \\
			\multicolumn{1}{l }{\hspace{0.1cm}+ 12-layer encoder (DLCL)} & 29.17 \\

			\bottomrule[1pt]
		\end{tabular}
		
		\vspace{-0.5em}
		\caption{Ablation results by tokenized BLEU [\%] on the test set of WMT English-German translation.}
		\label{table:ablation}
		\vspace{-1.em}
	\end{center}
\end{table}

\begin{figure*}[thb]
	\begin{center}
		
		\begin{tabular}{cc}
			%\hline
			
			\setlength{\tabcolsep}{1pt}
			\renewcommand\arraystretch{1.}
			
			\multirow{3}{*}
			{
				\subfloat
				{
					\includegraphics[width=0.5\linewidth,height=13.5em]{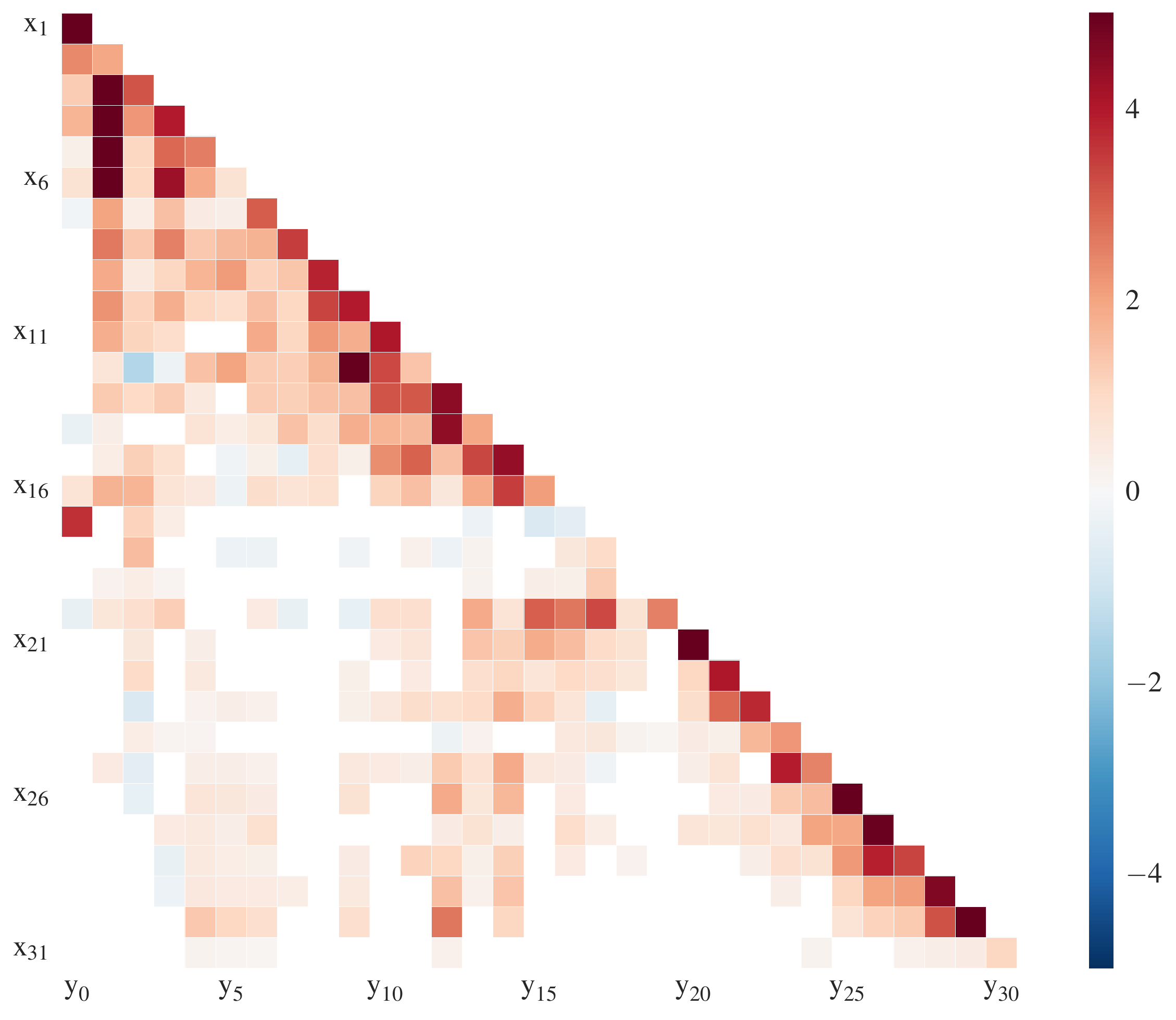}
					\label{fig:dla-encoder}
				}
			} & \\
			
			~ & \subfloat
			{
				\includegraphics[width=0.23\linewidth,height=4.7em]{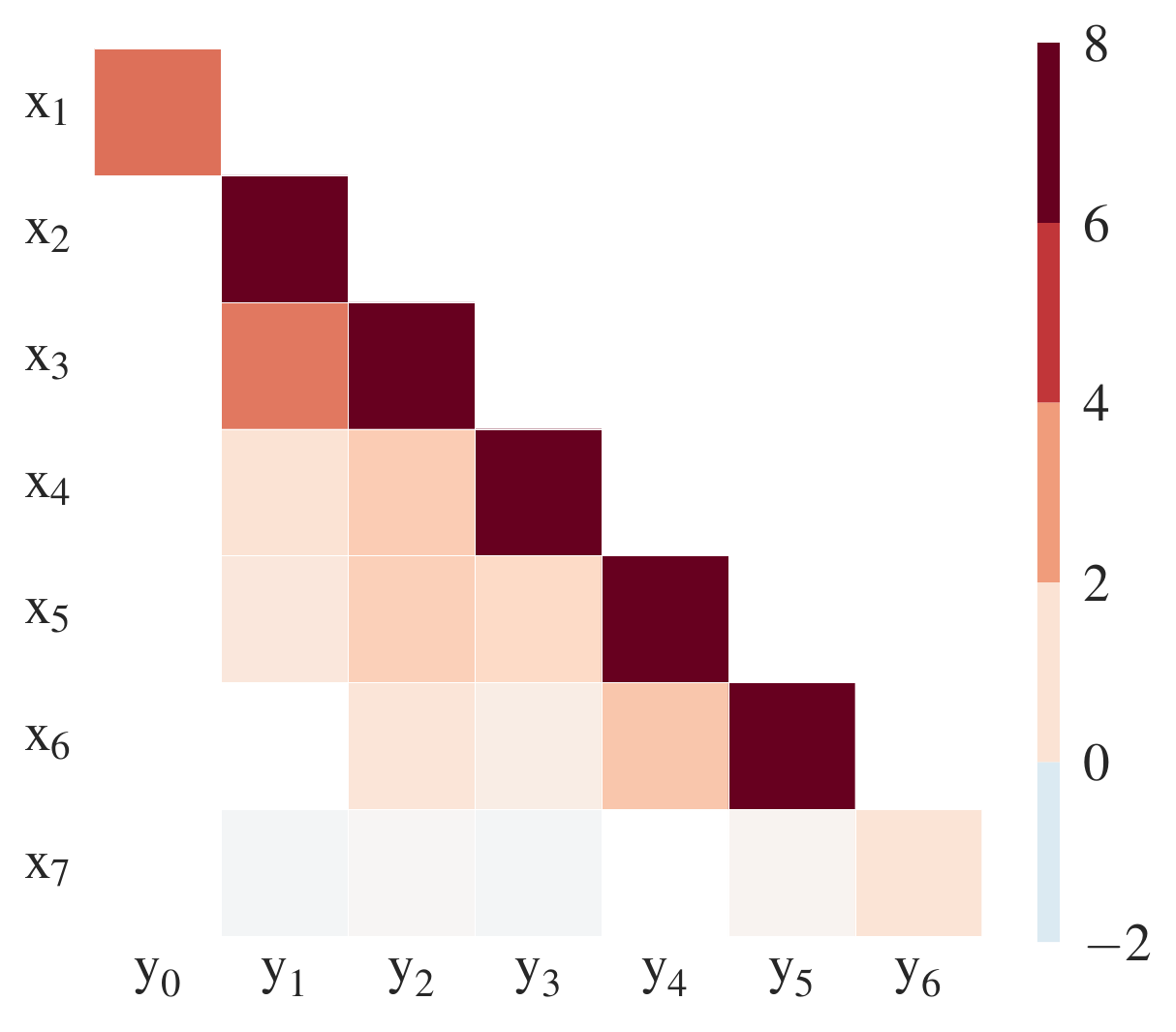}
				\label{fig:dla-decoder}
			} \\ %\hline
			~ & \footnotesize{(b) 6-layer decoder of \plainname} \\ %\hline
			
			~ & \subfloat
			{
				\resizebox {0.35\linewidth} {4.3em}
				{
					\begin{tikzpicture}
					\begin{scope}
					\setlength{\hnode}{\hnode}
					\tikzstyle{probnode} = [fill=blue!30,minimum width=0.8\hnode]
					\tikzstyle{labelnode} = [above]
					
					\node[] (w0) at (0,0) {};
					
					\node[probnode,anchor=south,minimum height=\hnode,inner sep=0.1pt,fill=ugreen!40,label=below:\scriptsize{$4.1$}] (attn11) at ([yshift=1\hnode]w0.north) {};
					
					\foreach \i / \j / \c in
					{11/12/3.3, 12/13/3.2, 13/14/1.7, 14/15/2.3,  15/16/1.1, 16/17/0.0,
						17/18/0.0, 18/19/0.1, 19/20/0.8, 20/21/0.5 }
					\node[probnode,anchor=south west,minimum height=(\c/4.1)*\hnode,inner sep=0.1pt,fill=ugreen!40,label=below:\scriptsize{\c}] (attn\j) at ([xshift=1pt]attn\i.south east) {};
					
					\node[anchor=south] (label1) at ([yshift=0.8em]attn17.north) {\textbf{\scriptsize{$x_{11} \sim x_{21}$}}};

					\node[probnode,anchor=north,minimum height=(0.2/4.1)*\hnode,inner sep=0.1pt,fill=ugreen!40,label=below:\scriptsize{$0.2$}] (attn22) at ([yshift=-0.3\hnode]w0.south) {};
					
					\foreach \i / \j / \c in
					{22/23/0.5, 23/24/0.0, 24/25/0.5, 25/26/0.2, 26/27/0.0, 27/28/0.0,
						28/29/0.1, 29/30/0.2, 30/31/0.0}
					{					
						\node[probnode,anchor=south west,minimum height=(\c/4.1)*\hnode,inner sep=0.1pt,fill=ugreen!40,label=below:\scriptsize{\c}] (attn\j) at ([xshift=1pt]attn\i.south east) {};
					}
					
					\node[anchor=south] (label2) at ([yshift=0.2em]attn28.north) {\scriptsize{ \textbf{$x_{22} \sim x_{31}$}}};
					
					\end{scope}
					\end{tikzpicture}	
				}	
				\label{fig:dla-indivisual}		
			} \\ %\hline
			
			\footnotesize{\rule{0pt}{4ex}  (a) 30-layer encoder of \plainname} & \footnotesize{(c) Weight distribution of $y_{10}$ in the encoder} \\[1pt]
			%\hline
			
		\end{tabular}

	\end{center}
	
	\begin{center}
		\vspace{-0.5em}
		\caption{ A visualization example of learned weights in our 30-layer pre-norm DLCL model.}
		\label{fig:learned_weights}
		\vspace{-1.em}
	\end{center}
\end{figure*}

\noindent We report the ablation study results in Table~\ref{table:ablation}. We first observe a modest decrease when removing the introduced layer normalization in Eq.~(\ref{eq-dla}). Then we try two methods to replace learnable weights with constant weights: \emph{All-One} ($W^i_j=1$) and \emph{Average} ($W^i_j=1/(i+1)$). We can see that these two methods consistently hurt performance, in particular in the case of All-One. It indicates that making the weights learnable is important for our model. Moreover, removing the added layer normalization in the Average model makes BLEU score drop by 0.28, which suggests that adding layer normalization helps more if we use the constant weights.
In addition, we did two interesting experiments on big models. The first one is to replace the base encoder with a big encoder in pre-norm Transformer-Base. The other one is to use DLCL to train a deep-and-wide Transformer (12 layers). Although both of them benefit from the increased network capacity, the gain is less than the ``thin'' counterpart in terms of BLEU, parameter size, and training efficiency.

\subsection{Visualization on Learned Weights}

\noindent We visually present the learned weights matrices of the 30-layer encoder (Figure~\ref{fig:learned_weights}(a)) and its 6-layer decoder (Figure~\ref{fig:learned_weights}(b)) in our pre-norm DLCL-30L model on En-De task. For a clearer contrast, we mask out the points with an absolute value of less than 0.1 or 5\% of the maximum per row. We can see that the connections in the early layers are dense, but become sparse as the depth increases. It indicates that making full use of earlier layers is necessary due to insufficient information at the beginning of the network. Also, we find that most of the large weight values concentrate on the right of the matrix, which indicates that the impact of the incoming layer is usually related to the distance between the outgoing layer. Moreover, for a fixed layer's output $y_i$, it is obvious that its contribution to successive layers changes dynamically (one column). To be clear, we extract the weights of $y_{10}$ in Figure~\ref{fig:learned_weights}(c). In contrast, in most previous paradigms of dense residual connection, the output of each layer remains fixed for subsequent layers.

\section{Related Work}

\begin{itemize}[leftmargin=*, label=]
	\item \textbf{Deep Models. }
	Deep models have been explored in the context of neural machine translation since the emergence of RNN-based models. To ease optimization, researchers tried to reduce the number of non-linear transitions \cite{zhou2016deep,wang2017deep}.
	But these attempts are limited to the RNN architecture and may not be straightforwardly applicable to the current Transformer model.
	Perhaps, the most relevant work to what is doing here is \citet{bapna2018training}'s work. They pointed out that vanilla Transformer was hard to train if the depth of the encoder was beyond 12. They successfully trained a 16-layer Transformer encoder by attending the combination of all encoder layers to the decoder.
	In their approach, the encoder layers are combined just after the encoding is completed, but not during the encoding process. In contrast, our approach allows the encoder layers to interact earlier, which has been proven to be effective in machine translation \cite{he2018layer} and text match \cite{lu2013deep}.
	In addition to machine translation, deep Transformer encoders are also used for language modeling \cite{devlin2018bert,al2018character}. For example, \citet{al2018character} trained a character language model with a 64-layer Transformer encoder by resorting to auxiliary losses in intermediate layers. This method is orthogonal to our \name method, though it is used for language modeling, which is not a very heavy task.
	
	\item \textbf{Densely Residual Connections. }
	Densely residual connections are not new in NMT. They have been studied for different architectures, e.g., RNN \cite{britz2017massive} and Transformer \cite{dou2018exploiting}. Some of the previous studies fix the weight of each layer to a constant, while others learn a weight distribution by using either the self-attention model \cite{wang2018multi} or a softmax-normalized learnable vector \cite{peters2018deep}. They focus more on learning connections from lower-level layers to the topmost layer. Instead, we introduce additional connectivity into the network and learn more densely connections for each layer in an end-to-end fashion.
\end{itemize}

\section{Conclusion}
\noindent We have studied deep encoders in Transformer. We have shown that the deep Transformer models can be easily optimized by proper use of layer normalization, and have explained the reason behind it. Moreover, we proposed an approach based on a dynamic linear combination of layers and successfully trained a 30-layer Transformer system. It is the deepest encoder used in NMT so far. Experimental results show that our thin-but-deep encoder can match or surpass the performance of Transformer-Big. Also, its model size is 1.6X smaller. In addition, it requires 3X fewer training epochs and is 10\% faster for inference.

\section*{Acknowledgements}
This work was supported in part by the National Natural Science Foundation of China (Grant Nos. 61876035, 61732005, 61432013 and 61672555), the Fundamental Research Funds for the Central Universities (Grant No. N181602013), the Joint Project of FDCT-NSFC (Grant No. 045/2017/AFJ), the MYRG from the University of Macau (Grant No. MYRG2017-00087-FST).

\bibliography{acl2019}
\bibliographystyle{acl_natbib}

\appendix

\section{Derivations of Post-Norm Transformer and Pre-Norm Transformer}
\label{sec:proof}

\noindent 
A general residual unit can be expressed by:
\begin{eqnarray}
\label{eq:app-y} y_l &=& x_l + \mathcal{F}(x_l; \theta_l), \\
\label{eq:app-x} x_{l+1} &=& f(y_l), 
\end{eqnarray}

\noindent where $x_l$ and $x_{l+1}$ are the input and output of the $l$-th sub-layer, and $y_l$ is the intermediate output followed by the post-processing function $f(\cdot)$.

We have known that the post-norm Transformer incorporates layer normalization ($\textrm{LN}(\cdot)$) by:
\begin{equation}
\label{eq:app-post}
\begin{split}
x_{l+1} & = \textrm{LN}\big(x_l + \mathcal{F}(x_l; \theta_l)\big) \\
& = \textrm{LN}\big(x_l + \mathcal{F}_{post}(x_l; \theta_l)\big)
\end{split}
\end{equation}
where $\mathcal{F}_{post}(\cdot)=\mathcal{F}(\cdot)$. Note that we omit the parameter in $\textrm{LN}$ for clarity. Similarly, the pre-norm Transformer can be described by:
\begin{equation}
\label{eq:app-prev}
\begin{split}
x_{l+1} &= x_l + \mathcal{F}\big(\textrm{LN}(x_l); \theta_l\big) \\
        &= x_l + \mathcal{F}_{pre}(x_l; \theta_l)
\end{split}
\end{equation}
where $\mathcal{F}_{pre}(\cdot)=\mathcal{F}(\textrm{LN}(\cdot))$. In this way, we can see that both post-norm and pre-norm are special cases of the general residual unit. Specifically, the post-norm Transformer is the special case when:
\begin{equation}
\label{eq:app-post-hf}
f_{post}(x)=\textrm{LN}(x), 
\end{equation}
while for pre-norm Transformer, it is:
\begin{equation}
\label{eq:app-prev-hf}
f_{pre}(x)=x. 
\end{equation}

Here we take a stack of $L$ sub-layers as an example. Let $\mathcal{E}$ be the loss used to measure how many errors occur in system prediction, and $x_L$ be the output of the top-most sub-layer. Then from the chain rule of back propagation we obtain:

\begin{equation}
\label{eq-bp-core}
\begin{split}
\frac{\partial{\mathcal{E}}}{\partial{x_{l}}} = \frac{\partial{\mathcal{E}}}{\partial{x_L}} \frac{\partial{x_L}}{\partial{x_{l}}}
\end{split}
\end{equation}

\noindent To analyze it, we can directly decompose $\frac{\partial{x_L}}{\partial{x_{l}}}$ layer by layer:
\begin{equation}
\label{eq-bp-core-sep}
\begin{split}
\frac{\partial{x_L}}{\partial{x_{l}}} = \frac{\partial{x_L}}{\partial{x_{L-1}}} \frac{\partial{x_{L-1}}}{\partial{x_{L-2}}} \ldots \frac{\partial{x_{l+1}}}{\partial{x_l}}. 
\end{split}
\end{equation}

\noindent Consider two adjacent layers as Eq.\ref{eq:app-y} and Eq.~\ref{eq:app-x}, we have:
\begin{equation}
\label{eq-bp-adjacent}
\begin{split}
\frac{\partial{x_{l+1}}}{\partial{x_l}} &= \frac{\partial{x_{l+1}}}{\partial{y_l}} \frac{\partial{y_l}}{\partial{x_l}} \\
&= \frac{\partial{f(y_l)}}{\partial{y_l}} \Big(1 + \frac{\partial{\mathcal{F}(x_l; \theta_l)}}{\partial{x_l}}\Big)
\end{split}
\end{equation}

For post-norm Transformer, it is easy to know $\frac{\partial{f_{post}(y_l)}}{\partial{y_l}}=\frac{\partial{\textrm{LN}(y_l)}}{\partial{y_l}}$ according to Eq.(\ref{eq:app-post-hf}). Then put Eq.(\ref{eq-bp-core-sep}) and (\ref{eq-bp-adjacent}) into Eq.(\ref{eq-bp-core}) and we can obtain the differential $\mathcal{L}$ w.r.t. $x_l$:
%\iffalse
\begin{eqnarray}
%\label{eq-post-bp-general}
\frac{\partial{\mathcal{E}}}{\partial{x_{l}}} & = & \frac{\partial{\mathcal{E}}}{\partial{x_L}} \times \prod_{k=l}^{L-1} \frac{\partial{\textrm{LN}(y_k)}}{\partial{y_k}} \times \nonumber \\
\label{eq-post-bp-general} & & \prod_{k=l}^{L-1} \Big(1 + \frac{\partial{\mathcal{F}(x_k; \theta_k)}}{\partial{x_k}}\Big)
\end{eqnarray}
%\fi

\noindent Eq.(\ref{eq-post-bp-general}) indicates that the number of product terms grows linearly with $L$, resulting in prone to gradient vanishing or explosion.

However, for pre-norm Transformer, instead of decomposing the gradient layer by layer in Eq.~(\ref{eq-bp-core-sep}), we can use the good nature that $x_L=x_l+\sum_{k=l}^{L-1} \mathcal{F}_{pre}(x_k; \theta_k)$ by recursively using Eq.~(\ref{eq:app-prev}):
\begin{equation}
\label{eq:app-pre-nature}
\begin{split}
x_L & = x_{L-1} + \mathcal{F}_{pre}(x_{L-1}; \theta_{L-1}) \\
& = x_{L-2} + \mathcal{F}_{pre}(x_{L-2}; \theta_{L-2}) + \mathcal{F}_{pre}(x_{L-1}; \theta_{L-1}) \\
& \cdots \\
& = x_l+\sum_{k=l}^{L-1} \mathcal{F}_{pre}(x_k; \theta_k)
\end{split}
\end{equation}

\noindent In this way, we can simplify Eq.(\ref{eq-bp-core-sep}) as:
\begin{equation}
\label{eq-prev-bp-core}
\begin{split}
\frac{\partial{x_L}}{\partial{x_{l}}} &= 1 + \sum_{k=l}^{L-1} \frac{\partial{\mathcal{F}_{pre}(x_k; \theta_k)}}{\partial{x_l}}
\end{split}
\end{equation}

\noindent Due to $\frac{\partial{f_{pre}(y_l)}}{\partial{y_l}}=1$, we can put Eq.~(\ref{eq-prev-bp-core}) into Eq.~(\ref{eq-bp-core}) and obtain:
\begin{equation}
\label{eq-prev-bp-general}
\begin{split}
\frac{\partial{\mathcal{E}}}{\partial{x_{l}}} &=
\frac{\partial{\mathcal{E}}}{\partial{x_L}} \times \Big(1 + \sum_{k=l}^{L-1} \frac{\partial{\mathcal{F}_{pre}(x_k; \theta_k)}}{\partial{x_l}}\Big) \\
&= \frac{\partial{\mathcal{E}}}{\partial{x_L}} \times \Big(1 + \sum_{k=l}^{L-1} \frac{\partial{\mathcal{F}(\textrm{LN}(x_k); \theta_k)}}{\partial{x_l}}\Big)
\end{split}
\end{equation}

\section{Training Hyper-parameters for Deep Models}
\label{sec:hyper-param}

\begin{table}[thb]
	\small
	\begin{center}
		\begin{tabular}{l | c c c c | c }
			%\hline
			%\hline
			\toprule[1pt]
			
			\multicolumn{1}{c |}{\textbf{Model}} &
			\multicolumn{1}{c }{\textbf{Batch}} &
			\multicolumn{1}{c }{\textbf{Upd.}} & 
			\multicolumn{1}{c }{\textbf{Lr}} &
			\multicolumn{1}{c |}{\textbf{Wu.}}&
			\multicolumn{1}{c }{\textbf{PPL}} \\
			\hline %\hline
			
			\multicolumn{1}{l|}{\footnotesize{post}} 		
			& 4096 & 100k & $7e^{-4}$ & 4k & 4.85 \\
			\multicolumn{1}{l|}{\footnotesize{post}}  	
			& 8192 & 50k & $2e^{-3}$	& 16k & * \\
			\cdashline{1-6}	
			
			\multicolumn{1}{l|}{\footnotesize{post-20L}}  	    
			& 4096 & 100k & $7e^{-4}$ & 4k & * \\
			\multicolumn{1}{l|}{\footnotesize{post-20L}}  	    
			& 8192 & 50k & $2e^{-3}$ & 16k & * \\
			%\cdashline{1-6}	
			\hline
			
			\multicolumn{1}{l|}{\footnotesize{pre}} 		
			& 4096 & 100k & $1e^{-3}$ & 8k & 4.88 \\
			\multicolumn{1}{l|}{\footnotesize{pre}}  	
			& 8192 & 50k & $2e^{-3}$	& 16k & {4.86} \\
			\cdashline{1-6}	
			
			\multicolumn{1}{l|}{\footnotesize{pre-20L}}  	    
			& 4096 & 100k & $1e^{-3}$ & 8k & 4.68 \\
			\multicolumn{1}{l|}{\footnotesize{pre-20L}}  	    
			& 8192 & 50k & $2e^{-3}$ & 16k & 4.60 \\
			
			\bottomrule[1pt]
		\end{tabular}
		
		\vspace{-0.0em}
		\caption{Hyper-parameter selection for shallow and deep models based on perplexity on validation set for English-German translation. ``post-20L'' is short for post-norm Transformer with a 20-layer encoder. Similarly, ``pre-20L'' denotes the pre-norm Transformer case. * indicates that the model failed to train.} 
		\label{table:hyper-param}
		\vspace{-0.5em}
	\end{center}
\end{table}

\noindent We select hyper-parameters by measuring perplexity on the validation set of WMT En-De task. We compare the effects of hyper-parameters in both shallow networks (6 layers) and deep networks (20 layers). We use the standard hyper-parameters for both models as the baselines. More concretely, for post-norm Transformer-Base, we set \textit{batch}/\textit{update}/\textit{lr}/\textit{warmup} to 4096/100k/7$\times$10$^{-4}$/4k as the original Transformer, while for pre-norm Transformer-Base, the configuration is 4096/100k/10$^{-3}$/8k as suggested in \emph{tensor2tensor}. As for deep models, we uniformly use the setting of 8192/50k/2$\times$10$^{-3}$/16k. Note that while we use a 2X larger batch size for deep models, we reduce a half of the number of updates. In this way, the amount of seen training data keeps the same in all experiments. A larger learning rate is used to speed up convergence when we use large batch. In addition, we found simultaneously increasing the learning rate and warmup steps worked best.

Table~\ref{table:hyper-param} report the results. First of all, we can see that post-norm Transformer failed to train when the network goes deeper. Worse still, the shallow network also failed to converge when switching to the setting of deep networks. We attribute it to post-norm Transformer being more sensitive to the large learning rate. On the contrary, in the case of either a 6-layer encoder or a 20-layer encoder, the pre-norm Transformer benefits from the larger batch and learning rate. However, the gain under deep networks is larger than that under shallow networks.

\end{document}